\documentclass[runningheads]{llncs}

\usepackage[T1]{fontenc}
\usepackage[utf8]{inputenc}
\usepackage{graphicx}
\usepackage{booktabs}
\usepackage{multirow}
\usepackage{tabularx}
\usepackage{array}
\usepackage{amsmath,amssymb,mathtools}
\usepackage{xcolor}
\usepackage{xspace}
\usepackage{enumitem}
\usepackage{float}
\usepackage{tikz}
\usetikzlibrary{arrows.meta,positioning,fit,calc}
\usepackage[hidelinks]{hyperref}
\usepackage[nameinlink,capitalise]{cleveref}

\definecolor{dynablue}{HTML}{284B63}
\definecolor{dynaorange}{HTML}{E07A5F}
\definecolor{dynagreen}{HTML}{3D9970}
\definecolor{dynapurple}{HTML}{725A9D}
\definecolor{todored}{HTML}{B42318}

\newcommand{\method}{DynaWM\xspace}
\newcommand{\benchmark}{\textsc{DynaGrasp-32}\xspace}
\newcommand{\dataset}{\textsc{DynaGrasp-1600}\xspace}

\newcommand{\R}{\mathbb{R}}
\newcolumntype{Y}{>{\centering\arraybackslash}X}
\title{\texorpdfstring{\method: A Base-VLA-Guided World Foundation Model for Moving-Object Manipulation}{DynaWM: A Base-VLA-Guided World Foundation Model for Moving-Object Manipulation}}
\titlerunning{\method for Moving-Object Manipulation}

\author{Chongkei Chang\inst{1}
\and Zhidong Deng\inst{1}\thanks{Corresponding author.}}
\authorrunning{C. Chang and Z. Deng}
\institute{Department of Computer Science and Technology, THUAI, Tsinghua University, Beijing 100084, China\\
\email{michael@tsinghua.edu.cn}}

\begin{document}
\maketitle

\begin{abstract}
% ===== BEGIN abstract/abstract_v7 =====
Although vision-language-action (VLA) models have received widespread attention, many challenges remain in manipulating dynamic moving objects.
In most existing approaches, end-to-end forward or inverse dynamics models, i.e., world models, are incorporated into high-performance base VLA architectures, which may degrade the performance of well-pretrained base VLA models due to inappropriate fine-tuning.
In this paper, we propose \method, a base-VLA-guided world foundation model that adapts to a wide variety of fine-tuned and coarse-tuned base-VLA checkpoints for moving-object manipulation.
\method uses a Mamba-3-based action encoder to encode the base action chunk produced by the base VLA into an action-conditioning representation, a V-JEPA~2.1 vision encoder to extract features from multi-view observation history, and a proprioceptive state encoder to encode robotic-arm proprioceptive states.
These feature representations jointly condition a flow-matching DiT to regenerate motion-aware action trajectories for moving-object manipulation.
For systematic evaluation, we construct the \benchmark benchmark, covering six categories of moving-object manipulation tasks, including velocity variation, trajectory variation, and multi-object manipulation, as well as the \dataset dataset, which consists of 32 scenarios, 1,600 demonstration trajectories, and approximately 1.53M images.
For fine-tuned base-VLA checkpoints, \method achieves percentage improvements of 7.19, 45.31, 1.88, and 10.94 over SmolVLA, X-VLA, $\pi_0$, and $\pi_{0.5}$, respectively.
For coarse-tuned base-VLA checkpoints, performance increases by 35.13, 44.06, 35.69, and 26.13 percentage, respectively.
Ablation experiments show that visual encoding enhances success by 27.50\%, while reducing success by 45.44\% if action conditioning is removed.

% ===== END abstract/abstract_v7 =====

\end{abstract}

% ===== BEGIN introduction/introduction_v7 =====
\section{Introduction}
\label{sec:introduction}

Vision-language-action (VLA) models connect visual--language knowledge with executable robot control.
RT-2 and RT-X transferred web-scale representations to robotic actions \cite{brohan2023rt,collaboration2023open}.
Octo and OpenVLA made generalist policies accessible \cite{team2024octo,kim2024openvla}.
Flow-based $\pi_0$ and $\pi_{0.5}$ improved continuous control and open-world adaptation \cite{black2024pi,intelligence2025pi}, while SmolVLA and X-VLA explored compact community models and cross-embodiment adaptation \cite{shukor2025smolvla,zheng2025x}.
Most recently, $\pi_{0.7}$ demonstrated a complementary route through a single steerable policy conditioned by language, metadata, history, and world-model-generated visual subgoals \cite{physicalintelligence2026pi07}.

Most progress, however, is achieved on tasks where the target is stationary or its motion is caused mainly by the robotic arm's own action.
Interception of independently moving targets remains underexplored, especially across heterogeneous VLA architectures.

Thus moving object grasping is a dynamic interception problem rather than a static manipulation one with extra appearance variation.
Throughout this paper, a base VLA refers to a VLA adapted to the benchmark and kept frozen during \method training and evaluation, while a base action chunk refers to the action sequence it emits.
Note that the base VLA predicts from current observations rather than visual history.
Such an observation is used to detect the position $p_t$ of the target instead of its future direction and speed.

When intercepting a moving target, the robotic arm cannot rely only on the currently observed position, because the target may have already moved during the delay from observation to action generation and execution.
Therefore, the model needs to use sequential multi-view history to predict the target's motion trend and continuously update the action trajectory during execution \cite{xie2026dynamicvla,fang2026towards,wei2026f2f,syed2026intercepting}.

Existing approaches attack different parts of this temporal gap.
Dynamics-aware VLAs such as DynamicVLA and PUMA jointly train perception and control for changing scenes \cite{xie2026dynamicvla,fang2026towards}.
World-model methods integrate future observations, latent states, or visual subgoals into a policy \cite{zhang2025dreamvla,sun2026vla,chen2026dial,tian2026starry,physicalintelligence2026pi07}.
AHEAD freezes OpenVLA but predicts future tokens in that specific model's representation space \cite{syed2026intercepting}.
Streaming and asynchronous action-generation methods reduce waiting and discontinuity \cite{jiang2025streaming,black2025real,lu2026faster}.
These works show that anticipation and low latency matter, but leave open whether one base-action-conditioned world foundation model can support multiple base-VLA checkpoints with different tuning extents.

Directly integrating a world-model configuration into each VLA is valid, but it ties motion modeling to model-specific visual tokens, action decoders, and training.
We instead use a world foundation model that takes action trajectories produced by base VLA policies such as SmolVLA, X-VLA, $\pi_0$, and $\pi_{0.5}$ as input and extracts action-conditioned representations.

Our key observation is that the emitted action chunk already reflects much of the semantic problem: \emph{what} to manipulate, the approaching direction, and the grasp phase.
Despite different VLA token spaces, output action trajectories share physical meaning and can serve as action-conditioning signals.
The missing signal is \emph{how the scene is changing}: sequential multi-view history lets the new module regenerate trajectories toward new interception points rather than chase the latest observed position.

The proposed \method is shown in Fig.~\ref{fig:system-workflow}.
For each base VLA, the unaugmented and augmented evaluations use the same frozen checkpoint.
As a result, any improvements cannot be attributed to base-VLA training.
A Mamba-3-based action encoder \cite{lahoti2026mamba} generates a representation of the sequential base action chunk, while a V-JEPA~2.1 encoder \cite{murlabadia2026v} represents observed multi-view history without rolling out future images or latent states.
Together with proprioceptive states, these conditions jointly condition a flow-matching DiT that generates the action chunk without accessing VLA language tokens, visual tokens, hidden states, or gradients.

Benchmarks such as Meta-World, LIBERO, and RoboTwin evaluate multi-task skill acquisition, knowledge transfer, long-horizon behavior, bimanual coordination, and scene variation \cite{yu2019meta,liu2023libero,mu2024robotwin,chen2025robotwin}.
Yet their variation usually comes from initial configurations, objects, instructions, or visual randomization.
Object motion is often induced by robotic-arm contact rather than treated as an independent pre-contact variable.
Dedicated dynamic-manipulation benchmarks establish the problem's importance \cite{burgesslimerick2022dgbench,liang2025whole}, but a common diagnostic suite for heterogeneous base VLA variants remains limited.

We therefore construct \benchmark, a 32-task benchmark with six diagnostic families: basic interception, appearance and geometry transfer, collision-driven path changes, visual variation, multi-object sequencing with language bindings, and speed variation.
As its paired dataset, our \dataset contains 1,600 demonstrations collected by human teleoperation in simulation, roughly 510K control steps, and 1.53M images from three views.
Each task is evaluated in 50 closed-loop episodes.

Our contributions are summarized below:
\begin{itemize}[leftmargin=*,nosep]
    \item We introduce \method, a base-VLA-guided world foundation model for moving-object manipulation. It adapts to various fine-tuned and coarse-tuned base-VLA checkpoints and uses the base action chunk outputted by each base VLA as a conditioning signal for regenerating actions toward moving targets without modifying the base-VLA architecture.
    \item We use the action chunk produced by the base VLA as a conditioning input and design an action-regeneration module composed of a Mamba-3-based action encoder, a V-JEPA~2.1 vision encoder, a proprioceptive state encoder, and a flow-matching DiT. The Mamba-3 encoder encodes the action chunk, the V-JEPA~2.1 encoder extracts features from multi-view observation history, and the proprioceptive state encoder represents robotic-arm states. These three types of features jointly condition the DiT to regenerate motion-aware execution trajectories.
    \item We construct both the \benchmark benchmark and the \dataset dataset for systematic evaluation of moving-object manipulation. They cover six dynamic task families, 32 scenarios, 1,600 demonstration trajectories, and approximately 1.53M images. Evaluation across four base VLAs and two checkpoint types, forming eight model-checkpoint combinations, shows consistent gains in model-level average success rates, ranging from $1.88\%$ to $45.31\%$.
\end{itemize}

% ===== END introduction/introduction_v7 =====
% ===== BEGIN related_work/related_work_v7 =====
\section{Related Work}
\label{sec:related}

\paragraph{Generalist VLA models.}
Robot learning has progressively moved from task-specific visuomotor policies toward generalist models that connect Internet-scale visual--language knowledge with physical control.
RT-2 transferred web knowledge into discretized robot actions \cite{brohan2023rt}, Open X-Embodiment and RT-X expanded training across robot embodiments \cite{collaboration2023open}, and Octo and OpenVLA made generalist policy development more accessible through open models and data recipes \cite{team2024octo,kim2024openvla}.
In parallel, continuous control shifted from direct regression toward diffusion and flow-based action generation \cite{chi2023diffusion,lipman2022flow}: RDT-1B applies diffusion to bimanual manipulation \cite{liu2024rdt}, while $\pi_0$ uses a flow-matching action expert and $\pi_{0.5}$ extends it toward open-world generalization \cite{black2024pi,intelligence2025pi}.
GR00T N1 targets generalist humanoid control, Gemini Robotics emphasizes embodied reasoning and cross-embodiment transfer, Qwen-VLA unifies manipulation, navigation, and trajectory prediction, and $\pi_{0.7}$ combines memory, multimodal prompting, and world-model-generated subgoals in a steerable policy \cite{nvidia2025gr00t,team2025gemini,wang2026qwen,physicalintelligence2026pi07}.
These systems pursue generality primarily through broader pretraining mixtures, specialized policy architectures, and model-specific adaptation.
They leave a complementary question: whether a base-action-conditioned world foundation model can provide missing capabilities without modifying an already trained base VLA or resuming its training.

\paragraph{Dynamic manipulation.}
Dynamic manipulation first developed around explicit interception and motion-aware control.
Early systems combined grasp reachability with target-motion prediction \cite{akinola2021dynamic} or learned dynamic grasping through adversarial reinforcement learning \cite{wu2022grasparl}.
DGBench then established a reproducible benchmark for grasping under relative motion \cite{burgesslimerick2022dgbench}; subsequent work extended the problem to whole-body legged manipulation \cite{liang2025whole}.
Modern VLA research replaces hand-designed motion pipelines with learned spatiotemporal policies.
DynamicVLA combines a compact model with continuous inference and action streaming \cite{xie2026dynamicvla}.
DOMINO provides large-scale dynamic training data, while PUMA combines historical optical flow with short-horizon future queries \cite{fang2026towards}; ST-VLA and LaMP introduce 4D-aware or scene-flow motion priors \cite{wu2026st,wang2026lamp}, and OFlow injects object-aware temporal flow into action generation \cite{wang2026oflow}.
F2F-AP predicts future observations to compensate for asynchronous-policy latency \cite{wei2026f2f}.
Most closely related, AHEAD wraps a frozen OpenVLA with a latent predictive world model \cite{syed2026intercepting}.
Together, these works establish the value of temporal prediction, but they generally either train a dedicated dynamic policy or operate inside a particular policy's feature space.
However, base-action-conditioned generation without access to policy-specific internal representations remains underexplored across structurally different frozen base VLAs.

\paragraph{Time-Series Prediction and Action Refinement.}
Two complementary lines of work address the mismatch between an observed scene and the action eventually executed.
The first predicts how the world will evolve.
Cosmos and Cosmos~3 extend world foundation models toward Physical AI \cite{nvidia2025cosmos,nvidia2026cosmos3}.
Generative world models couple video and action diffusion or use generated visual futures to guide control \cite{zhu2025unified,zhang2025dreamvla,ma2026dit4dit}, while latent approaches predict action-relevant state transitions or structured temporal features \cite{sun2026atomvla,chen2026dial,tian2026starry}.
V-JEPA and V-JEPA~2 learn predictive video representations without pixel-level reconstruction \cite{bardes2024revisiting,assran2025v}, and V-JEPA~2.1 exposes denser spatiotemporal features \cite{murlabadia2026v}.
The second line revises when and how action chunks are generated.
Action chunking improves temporal coherence and amortizes policy inference \cite{zhao2023learning}, but an open-loop chunk can become stale as the scene changes.
Streaming Flow Policy aligns flow progression with execution time \cite{jiang2025streaming}; Real-Time Chunking asynchronously inpaints the next chunk without retraining the underlying policy \cite{black2025real}; and FASTER prioritizes near-term actions to reduce reaction latency \cite{lu2026faster}.
Native continuation and adaptive chunking study cross-cycle reuse and when to replan \cite{liu2026learning,feng2026denoising}, while DFM-VLA and $\pi_0$-EqM enable iterative action correction \cite{chen2026dfm,liu20260}.
Some policy-internal methods instead maintain short-horizon action context or action-updated scene state inside the policy \cite{lian2026intentvla,zhang2026evoscene}.
\method connects these two lines through a world foundation model conditioned on base action chunks, recent multi-view history, and robotic-arm state.
A Mamba-3-based action encoder \cite{gu2023mamba,lahoti2026mamba} preserves the base chunk's sequential structure, while visual motion features provide scene-change evidence for regeneration without accessing policy-specific hidden states.

\paragraph{Benchmarks and datasets.}
Manipulation benchmarks have expanded along several axes.
RLBench and Meta-World established multi-task evaluation \cite{james2019rlbench,yu2019meta}; CALVIN emphasized language-conditioned long-horizon control \cite{mees2021calvin}; and LIBERO studied lifelong knowledge transfer \cite{liu2023libero}.
RoboCasa and VLABench broaden scene and long-horizon diversity \cite{nasiriany2024robocasa,zhang2024vlabench}.
RoboTwin and RoboTwin~2.0 scale bimanual task generation and domain randomization \cite{mu2024robotwin,chen2025robotwin}.
Large datasets such as BridgeData~V2, MimicGen, and DROID provide increasingly diverse demonstrations for policy training \cite{walke2023bridgedata,mandlekar2023mimicgen,khazatsky2024droid}.
Dynamic benchmarks form a smaller but growing branch: DGBench isolates relative-motion grasping, whole-body legged manipulation extends dynamic grasping across embodiments, and DOMINO introduces a broad dynamic task hierarchy \cite{burgesslimerick2022dgbench,liang2025whole,fang2026towards}.
What remains limited is a diagnostic suite that controls distinct sources of motion difficulty while supporting the same closed-loop comparison across heterogeneous base-VLA checkpoints with different tuning extents.
\benchmark targets this gap through six task families: basic interception, transfer across object appearance and geometry, collision-driven trajectory change, visual variation, multi-object sequencing with language bindings, and controlled variation in target velocity.

% ===== END related_work/related_work_v7 =====
% ===== BEGIN model/model_v7 =====
\section{Method}
\label{sec:method}

\subsection{Problem Formulation}

At time $t$, a base VLA $\pi_b$ receives the current observation $o_t$, language instruction or subtask $\ell_t$, and robotic-arm state $s_t$. It predicts a base action chunk
\begin{equation}
    A_t^b = \pi_b(o_t,\ell_t,s_t)
    \in \R^{H\times D_a},
\end{equation}
where $\ell_t$ denotes the instruction or subtask at time $t$, $H$ is the action horizon, and $D_a$ is the action dimension.
In our implementation, $H=10$ and $D_a=7$.
The base-VLA checkpoint is fixed.
It is not updated during \method training.
Our goal is to learn \method as a conditional world foundation model $g_\theta$ with trainable parameters $\theta$ that produces a motion-aware action chunk
\begin{equation}
    \hat A_t = g_\theta(A_t^b,V_t,s_t),
    \label{eq:dynawm}
\end{equation}
where $V_t$ is a short multi-view video ending at $t$.
\method does not access the base VLA's language, visual, or action tokens; it uses emitted physical actions as one condition, making the interface compatible with VLAs whose internal architectures and representation dimensions differ.

\subsection{Design Principle: From Reaction to Interception}

The base VLA observes a current scene and proposes how to act from that scene.
For a stationary target, executing this proposal later usually remains spatially valid; for a moving target, aiming at the current position becomes reactive tracking, causing the robotic arm to follow where the object was observed rather than where the robotic arm and object can meet.
Let $p_t$ and $p_{t+1}$ denote target positions in two consecutive observations.
The local target displacement is
\begin{equation}
    \delta_t = p_{t+1}-p_t,
\end{equation}
which cannot generally be identified from the current observation $o_t$ alone.
Recent frames provide temporal differences that constrain $\delta_t$ through direction, relative speed, and path-change evidence.

\method does not explicitly regress $p_t$, $p_{t+1}$, velocity, or $\delta_t$.
Instead, it learns the operational equivalent: visual motion features condition a new action trajectory whose contact region and timing shift away from the base VLA's current-position response.
We decompose the required information into three conditions:
\begin{equation}
    \begin{aligned}
        A_t^b &: \text{base action chunk},\\
        V_t   &: \text{recent scene motion},\\
        s_t   &: \text{current robotic-arm state}.
    \end{aligned}
\end{equation}

\method recombines these three inputs without reinterpreting language or requiring the base VLA to relearn motion estimation.

\begin{figure*}[t]
\centering
\includegraphics[width=0.98\textwidth]{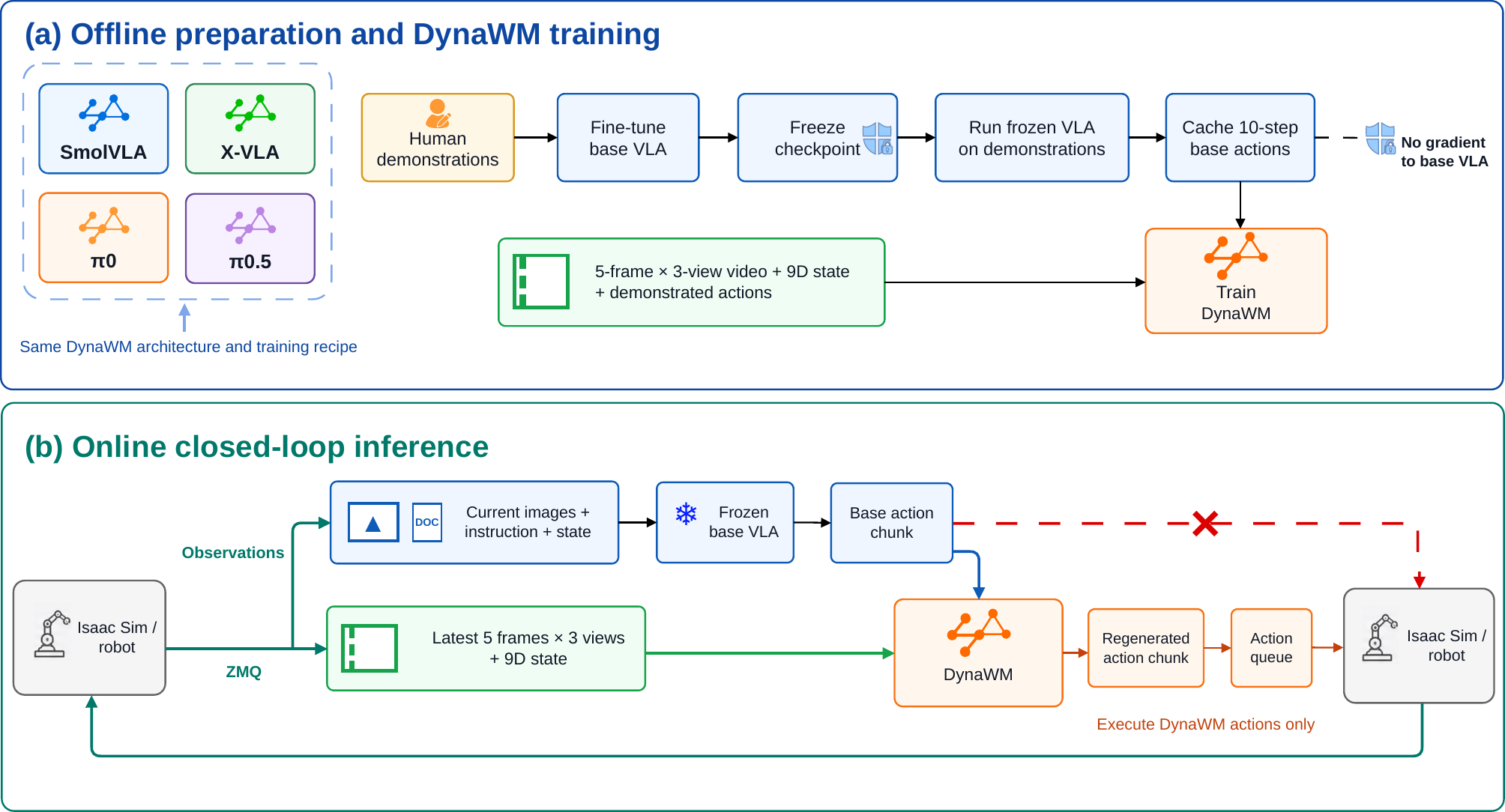}
\caption{\textbf{System framework.}
(Offline) The frozen base VLA is run over demonstrations to cache base action chunks, which train the corresponding \method model together with recent multi-view history, robotic-arm states, and demonstrated actions.
(Online) \method combines the base action chunk with recent visual dynamics to regenerate a motion-aware chunk for execution.
The base action chunk is used as a condition rather than executed directly, and the \method architecture and training recipe are shared across the four base VLA variants.}
\label{fig:system-workflow}
\end{figure*}

Fig.~\ref{fig:system-workflow} summarizes offline training and online execution.
Offline, \method trains from cached base action chunks, recent multi-view history, robotic-arm states, and demonstrated actions.
Online, the base action chunk is used only as a condition for regenerating a motion-aware chunk.
The unaugmented and augmented evaluations use the same frozen base-VLA checkpoint, separating action-conditioned generation from base-VLA training.
The following subsections define the action, visual, and generative components of \method.

\begin{figure*}[t]
\centering
\includegraphics[width=0.98\textwidth]{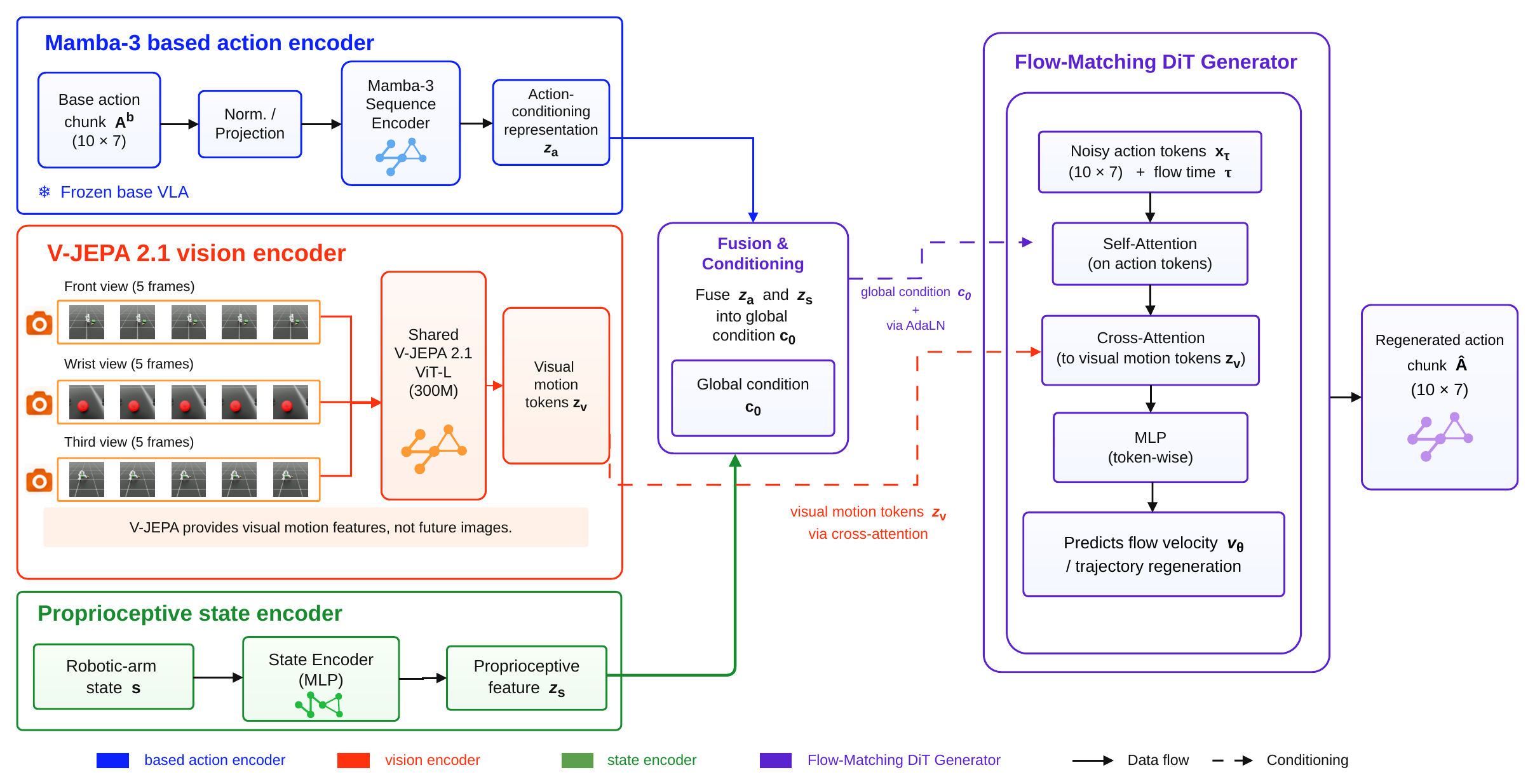}
\caption{\textbf{Internal architecture of \method.}
Mamba-3 encodes the base action chunk into an action-conditioning representation, and V-JEPA~2.1 encodes recent multi-view observations into visual motion tokens.
Action and robotic-arm-state features modulate the DiT through adaptive LayerNorm, visual motion tokens provide motion evidence through cross-attention, and the conditional flow generates the action chunk from Gaussian noise.}
\label{fig:dynawm-architecture}
\end{figure*}

\subsection{Action Trajectories as Conditioning Signals}

Different VLAs may use autoregressive tokens, diffusion, flow matching, or separate action experts, but when instantiated on the same robotic-arm interface their output action trajectories share physical meaning.
We treat $A_t^b$ as an observable short-horizon conditioning signal: its sequential structure captures approach direction, gripper phase, local control trend, and transitions between reaching, grasping, and transport.
Because adjacent actions are coupled by control continuity, we encode the chunk with a state-space sequence model whose recurrent scan accumulates trajectory context in action order.
Mamba-3 provides an efficient encoder for this continuous action signal, while alternative temporal encoders could be studied within the same framework.

Actions are normalized per dimension using statistics from cached base action chunks and demonstrations.
Let $E_a$ denote the Mamba-3 action encoder.
Mamba-3 processes the normalized sequence $\bar A_t^b$ in trajectory order:
\begin{equation}
    z_a = E_a(\bar A_t^b).
\end{equation}

In our implementation, the ordered chunk is mapped through a 128-dimensional Mamba state to a 256-dimensional action-conditioning representation.
The representation at the final valid step summarizes the preceding control sequence and serves as the action-conditioning representation.

The output actions do not require recovering all language or visual information inside the base VLA.
They only need to carry enough task-relevant action information to condition dynamic execution.
\Cref{sec:ablations} tests whether this restricted conditioning signal becomes a performance bottleneck.

\subsection{Multi-View Short-Horizon Dynamics}

For each synchronized camera view $k$, we collect a recent clip
\begin{equation}
    V_t^k = \{I_{t-L+1}^k,\ldots,I_t^k\},
\end{equation}
where $I_t^k$ is the image from view $k$ at time $t$ and $L$ denotes the observation-history length.
In our implementation, $L=5$.
Each clip is encoded by a shared V-JEPA~2.1 model \cite{murlabadia2026v}.
During \method inference, each observed clip is mapped directly to dense spatiotemporal tokens.
No future image or latent-state rollout is performed:
\begin{equation}
    Z_v =
    [E_v(V_t^1);\ldots;E_v(V_t^K)],
\end{equation}
where $E_v$ denotes the V-JEPA~2.1 vision encoder and $K$ is the number of camera views.
In our implementation, $K=3$.
The dense tokens preserve local spatial structure and temporal consistency.
Inter-frame displacement and its changes indicate direction, relative speed, rebound, or acceleration, while multiple views reduce ambiguity under wrist motion and partial occlusion.
These cues let the generator learn motion-dependent interception behavior without explicit velocity or future-coordinate labels.
The V-JEPA encoder is part of the trainable \method model; only the base VLA is frozen.

\subsection{Conditional Flow-Based Trajectory Regeneration}

Fig.~\ref{fig:dynawm-architecture} shows the internal architecture of \method.
Our generator is called a Diffusion Transformer (DiT), operating on action tokens.
The current robotic-arm state is embedded as $z_s=E_s(s_t)$, where $E_s$ is the proprioceptive state encoder, and concatenated with the action representation into the global condition $c_0=[z_a;z_s]$.
Visual motion tokens remain unpooled and enter the generator through cross-attention, allowing action tokens to attend to local motion evidence.
The DiT uses eight transformer blocks with 512-dimensional hidden states and eight attention heads.
Each block contains action-token self-attention, cross-attention to $Z_v$, and a token-wise MLP, with adaptive LayerNorm modulation from $c_0$ and the flow-time embedding.
Thus, the action condition persists throughout generation and visual motion evidence can revise each action token.

Let the normalized demonstration action chunk be $x_1=\bar A_t^{\mathrm{gt}}$ and let $x_0=\epsilon$, where $\epsilon\sim\mathcal{N}(0,I)$.
For a sampled flow time $\tau\in[0,1]$, we construct the linear probability path
\begin{equation}
    x_\tau=\tau x_1+(1-\tau)x_0,
    \qquad
    u_\tau=x_1-x_0.
\end{equation}

The conditional velocity field is trained with
\begin{equation}
    \mathcal{L}_{\mathrm{FM}}
    =
    \mathbb{E}\left[
    \left\|
    v_\theta(x_\tau,\tau\mid z_a,z_s,Z_v)-u_\tau
    \right\|_2^2
    \right].
    \label{eq:flow-loss}
\end{equation}

This is called trajectory \emph{regeneration}: the base action chunk is a condition but not the initial sample, and the output is not constrained to the form $A_t^b+\Delta A$.
Starting from noise permits larger corrections when a reactive base action chunk points toward an outdated target location; fixed residuals or flows initialized at that chunk may bias learning toward local repair, whereas regeneration can preserve the base action information while reconstructing interception geometry.
At inference, we draw an initial sample from $\mathcal{N}(0,I)$, numerically integrate the learned velocity field with ten Euler steps, and de-normalize the resulting chunk into robotic-arm action units.

For training, each base VLA is run once over the demonstrations to cache base action chunks; \method then trains from cached base action chunks, recent multi-view clips, robotic-arm states, and demonstrated chunks, with no base VLA in the training graph.
Separate \method weights fit each base VLA's action distribution, while the architecture and optimization recipe remain fixed across SmolVLA, X-VLA, $\pi_0$, and $\pi_{0.5}$.
At deployment, \method receives action chunks emitted by the frozen base VLA; both evaluations use the same base-VLA checkpoint to separate action-conditioned generation from base-VLA training.

% ===== END model/model_v7 =====
% ===== BEGIN experiments/experiments_v7 =====
\section{Experiments}
\label{sec:experiments}

\subsection{Benchmark and Evaluation Protocol}
\label{sec:benchmark}

\begin{figure}[H]
\centering
\includegraphics[width=0.99\textwidth]{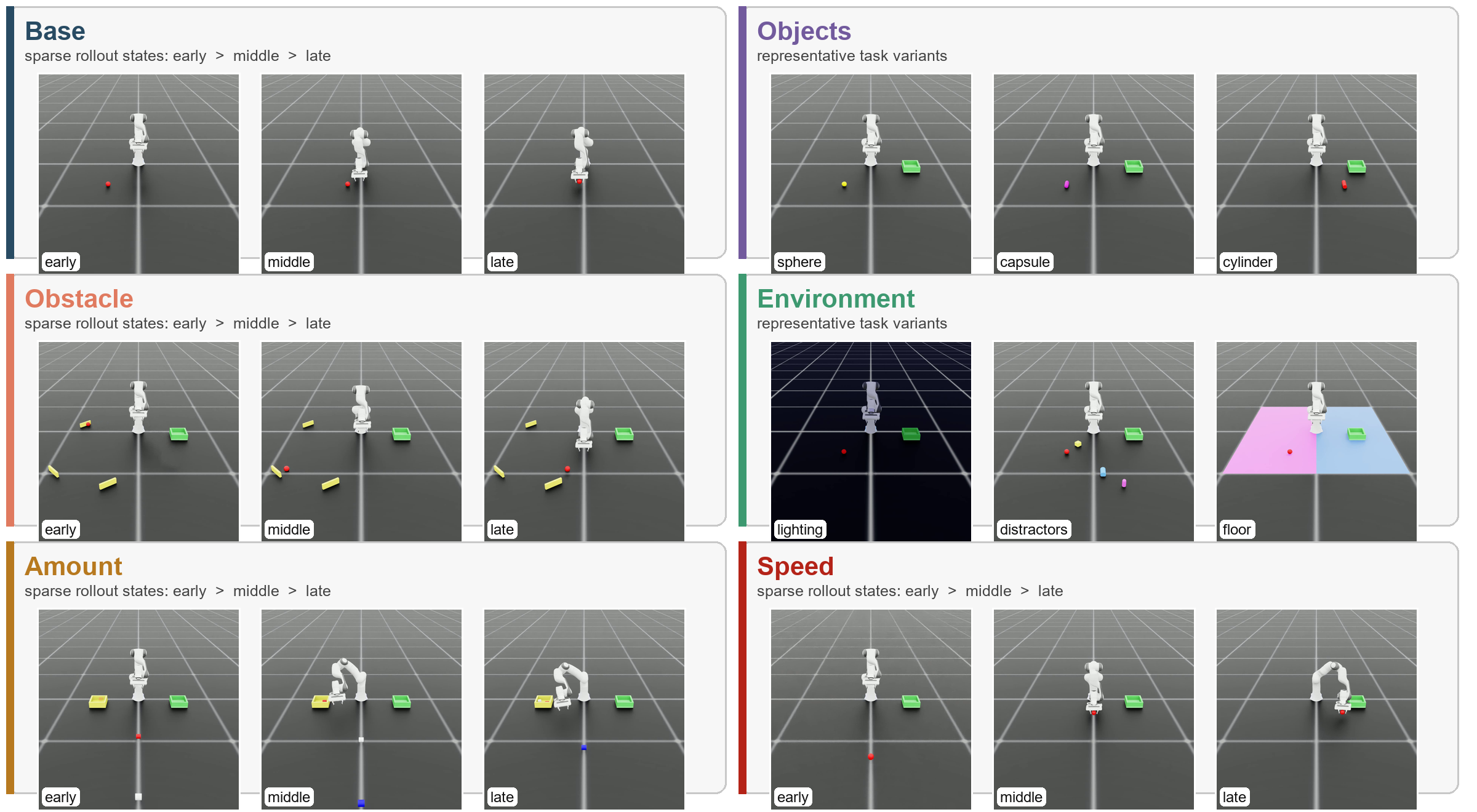}
\caption{\textbf{Simulation platform and diagnostic structure of \benchmark.}
All panels use the same Franka Panda workspace and front-camera pose.
Objects and Environment show representative variants; Base, Obstacle, Amount, and Speed show temporally separated rollout states.
Frame indices indicate time, not predicted object trajectories.}
\label{fig:benchmark-overview}
\end{figure}

\benchmark contains 32 Isaac Sim tasks in which the target object moves during physical action execution.
All tasks are grouped into six diagnostic families, categorized as appearance transfer, collision response, scene variation, multi-object sequencing, velocity variation, and basic interception (\cref{tab:task-families}).
Fig.~\ref{fig:benchmark-overview} shows representative layouts and rollout states.

\begin{table}[H]
\centering
\caption{The six diagnostic families of \benchmark.}
\label{tab:task-families}
\small
\setlength{\tabcolsep}{4pt}
\begin{tabularx}{\columnwidth}{l c X}
\toprule
Family & \# & Primary variation and capability \\
\midrule
Base & 2 & Orthogonal motion directions; basic interception \\
Objects & 6 & Color, shape, size, and direction; object transfer \\
Obstacle & 6 & Collision surfaces and rebound; path-change response \\
Environment & 6 & Illumination, floor, and distractors; visual robustness \\
Amount & 6 & Multiple objects and container bindings; sequencing \\
Speed & 6 & Episode-wise velocity variation; motion and timing \\
\bottomrule
\end{tabularx}
\end{table}

The benchmark uses the Franka Panda arm, three $256\times256$ RGB cameras, a 9D robotic-arm state, a 7D action space, and a 30\,Hz environment.
In \cref{tab:task-families}, the Base tasks require lifting a moving ball to at least 0.20\,m, while the remaining tasks generally require placing moving objects into language-specified baskets; multi-object tasks require all objects to be placed correctly.
The six families evaluate basic interception (Base), appearance and geometry transfer (Objects), collision-induced path changes (Obstacle), visual robustness under scene variation (Environment), multi-object interception with object--container binding and ordering (Amount), and episode-wise velocity variation (Speed).

\paragraph{Simulation platform and reproducibility.}
We use Isaac Sim because the benchmark requires articulated robotic-arm control, rigid-body contact and rebound, synchronized multi-view sensing, and controlled intervention on motion, appearance, and scene geometry within one closed-loop environment.
Isaac Sim provides physics, sensing, randomization, synthetic-data tools, and ROS~2 interfaces \cite{nvidia2026isaacsim}, letting \benchmark vary one dynamic difficulty source while holding the robotic arm, control frequency, observations, and success criteria fixed.
Its digital specification supports reproducibility through distributable scene assets, task logic, motion distributions, camera geometry, control timing, and success conditions.
We therefore position \benchmark as a shared diagnostic testbed that supports reproducible evaluation and provides a controlled foundation for subsequent hardware validation.

\dataset provides 50 demonstrations per task, collected by human teleoperation in simulation, totaling 1,600 trajectories, approximately 510K action steps, and 1.53M RGB images across three cameras.
We evaluate SmolVLA \cite{shukor2025smolvla}, X-VLA \cite{zheng2025x}, $\pi_0$ \cite{black2024pi}, and $\pi_{0.5}$ \cite{intelligence2025pi} using fine-tuned checkpoints (4--5 epochs for SmolVLA/X-VLA and 2--3 for $\pi_0/\pi_{0.5}$) and coarse-tuned checkpoints (about 0.5 epoch).
Every model--task configuration is evaluated in closed loop for 50 episodes, so the 32-task macro-average equals the pooled success rate over 1,600 episodes.
Augmented and unaugmented variants use the same frozen base-VLA checkpoint; the \method architecture and optimization recipe are fixed across base VLA variants, with separate \method weights for each base VLA's action distribution.
Experiments are run on an NVIDIA RTX PRO 6000 GPU with 96 GB memory.

\subsection{Dynamic Capability Across VLAs}

\begin{table*}[t]
\centering
\caption{\textbf{The success rate on \benchmark.}
Each value averages 32 tasks with 50 episodes per task.
Up/Tie/Down counts task-level changes after adding \method.
Gains are computed from unrounded success rates.}
\label{tab:main-results}
\small
\setlength{\tabcolsep}{8pt}
\resizebox{\textwidth}{!}{%
\begin{tabular}{llrrrr}
\toprule
Checkpoint & Base VLA & Base SR (\%) & + \method (\%) & Gain (\%) & Up/Tie/Down \\
\midrule
\multirow{4}{*}{Fine-tuned}
& SmolVLA & 62.50 & 69.69 & +7.19 & 23/1/8 \\
& X-VLA & 36.56 & \textbf{81.88} & \textbf{+45.31} & 32/0/0 \\
& $\pi_0$ & 72.56 & 74.44 & +1.88 & 14/4/14 \\
& $\pi_{0.5}$ & 64.63 & 75.56 & +10.94 & 21/5/6 \\
\midrule
\multirow{4}{*}{Coarse-tuned}
& SmolVLA & 40.63 & 75.75 & +35.13 & 32/0/0 \\
& X-VLA & 37.69 & \textbf{81.75} & \textbf{+44.06} & 32/0/0 \\
& $\pi_0$ & 39.88 & 75.56 & +35.69 & 31/1/0 \\
& $\pi_{0.5}$ & 48.44 & 74.56 & +26.13 & 31/0/1 \\
\bottomrule
\end{tabular}
}
\end{table*}

\Cref{tab:main-results} reports the results achieved in the primary cross-VLA evaluation.
\method improves the 32-task average in all eight combinations of base VLA and checkpoint type while leaving every base-VLA checkpoint unchanged.
For fine-tuned checkpoints, the equal-weight mean rises from $59.06\%$ to $75.39\%$ ($+16.33\%$), and for coarse-tuned checkpoints from $41.66\%$ to $76.91\%$ ($+35.25\%$).
Across all eight comparisons, mean success increases from $50.36\%$ to $76.15\%$.

The gain is largest when the base VLA has more dynamic headroom.
Fine-tuned $\pi_0$ begins at $72.56\%$ and gains only $1.88\%$, whereas fine-tuned X-VLA begins at $36.56\%$ and gains $45.31\%$.
For every base VLA variant, coarse-tuned checkpoints also yield a larger gain than fine-tuned checkpoints.
After regeneration, the eight systems occupy a narrower $69.69\%$--$81.88\%$ range, supporting a capability-compensation interpretation.
Task-level counts show the boundary: the four coarse-tuned checkpoints improve on at least 31 of 32 tasks, but the already strong fine-tuned $\pi_0$ improves on 14 tasks and regresses on 14.
Its clearest regressions occur in multi-object Amount tasks, suggesting that local motion correction can disturb an otherwise valid object order or language-conditioned binding.

\subsection{Where Do the Gains Come From?}

\begin{table*}[t]
\centering
\caption{\textbf{The success rate across diagnostic families.}
Values are averaged within each task family and then across the four base VLA variants; global 32-task averages remain the primary metric.}
\label{tab:family-results}
\small
\setlength{\tabcolsep}{7pt}
\resizebox{\textwidth}{!}{%
\begin{tabular}{lrrrrrr}
\toprule
& \multicolumn{3}{c}{Fine-tuned checkpoints} & \multicolumn{3}{c}{Coarse-tuned checkpoints} \\
\cmidrule(lr){2-4}\cmidrule(lr){5-7}
Family & Base & + \method & Gain & Base & + \method & Gain \\
\midrule
Base & 66.25 & 88.25 & +22.00 & 49.75 & 83.75 & +34.00 \\
Objects & 80.17 & 91.83 & +11.67 & 65.83 & 92.42 & +26.58 \\
Obstacle & 43.17 & 65.25 & +22.08 & 23.08 & 68.83 & +45.75 \\
Environment & 73.17 & 85.67 & +12.50 & 59.75 & 84.75 & +25.00 \\
Amount & 48.17 & 61.00 & +12.83 & 26.25 & 62.08 & +35.83 \\
Speed & 48.25 & 68.92 & +20.67 & 30.67 & 74.17 & +43.50 \\
\bottomrule
\end{tabular}
}
\end{table*}

\Cref{tab:family-results} shows that the aggregate improvement is not driven by one easy task family.
For fine-tuned checkpoints, the largest mean gains occur on Obstacle ($+22.08\%$), Base ($+22.00\%$), and Speed ($+20.67\%$).
For coarse-tuned checkpoints, Obstacle and Speed improve by $45.75\%$ and $43.50\%$.
These settings require updating contact geometry from motion evidence because velocity varies across episodes or collisions invalidate pre-impact trajectories.

Objects and Environment also improve, showing that the learned motion-conditioned generation remains useful under appearance and scene variation, but Amount exposes the clearest boundary.
It improves weak policies yet can regress for strong ones: fine-tuned $\pi_{0.5}$ drops $4.33\%$ on Amount despite an overall gain of $10.94\%$.
These tasks require object--container bindings and completion order across chunks, so local interception alone does not ensure correct sequencing.
\subsection{What Makes Action-Conditioned Generation Work?}
\label{sec:ablations}

\subsubsection{Temporal Visual Evidence}

\begin{table*}[t]
\centering
\caption{\textbf{The ablation on visual dynamics.}
All values are success rates averaged over the indicated task family.
These systems use V-JEPA features to augment the action model, but do not use the base action chunk as a conditioning input.}
\label{tab:visual-ablation}
\small
\setlength{\tabcolsep}{5.5pt}
\resizebox{\textwidth}{!}{%
\begin{tabular}{lrrrrrrr}
\toprule
Configuration & All & Base & Objects & Obstacle & Environment & Amount & Speed \\
\midrule
Qwen3.5-0.8B & 39.94 & 33.00 & 72.33 & 25.67 & 27.67 & 60.33 & 16.00 \\
\quad + V-JEPA~2.1 80M & 55.00 & 68.00 & 79.67 & 47.33 & 61.33 & 38.00 & 44.33 \\
\quad + V-JEPA~2.1 300M & \textbf{67.44} & \textbf{77.00} & \textbf{91.00} & \textbf{56.67} & \textbf{75.00} & 54.33 & \textbf{57.00} \\
\bottomrule
\end{tabular}
}
\end{table*}

\Cref{tab:visual-ablation} shows that sequential visual features substantially improve the image--language action model: V-JEPA~2.1 80M raises average success from $39.94\%$ to $55.00\%$, and the 300M encoder reaches $67.44\%$.
The 300M model improves every family over 80M, indicating that richer observation history provides useful motion evidence for dynamic action generation.

\subsubsection{Base Action Chunks as Conditioning Signals}

\begin{table*}[t]
\centering
\caption{\textbf{Base-action conditioning diagnostics.}
(a) The Qwen--V-JEPA reference uses language-conditioned visual features inside the model, whereas \method uses the base action chunk as its base-policy condition.
(b) The direct-start baseline tests whether the base action chunk is more effective as flow initialization or as a persistent encoded condition.}
\label{tab:interface-diagnostics}
\small
\begin{minipage}[t]{0.52\textwidth}
\centering
\textbf{(a) Base-action conditioning diagnostic}\\[2pt]
\setlength{\tabcolsep}{5pt}
\begin{tabular}{lr}
\toprule
Configuration & SR (\%) \\
\midrule
Internal Qwen + V-JEPA 300M & 67.44 \\
\midrule
SmolVLA + \method & 69.69 \\
X-VLA + \method & 81.88 \\
$\pi_0$ + \method & 74.44 \\
$\pi_{0.5}$ + \method & 75.56 \\
\midrule
\method mean & 75.39 \\
\bottomrule
\end{tabular}
\end{minipage}\hfill
\begin{minipage}[t]{0.42\textwidth}
\centering
\textbf{(b) Action-conditioning formulation}\\[2pt]
\setlength{\tabcolsep}{5pt}
\begin{tabularx}{\linewidth}{Xr}
\toprule
Configuration & SR (\%) \\
\midrule
Frozen $\pi_0$ & 72.56 \\
Direct action start, no action encoder & 29.00 \\
$\pi_0$ + \method & \textbf{74.44} \\
\bottomrule
\end{tabularx}
\end{minipage}
\end{table*}

As a reference, the Qwen--V-JEPA model directly fuses language-conditioned image features with temporal visual features inside the model, whereas \method uses only the base action chunk as a conditioning input.
Nevertheless, all four fine-tuned \method variants exceed the $67.44\%$ internal reference, with a mean of $75.39\%$ (\cref{tab:interface-diagnostics}(a)), suggesting that the action sequence retains enough short-horizon action information for action-conditioned generation.

\subsubsection{Why Encode the Action Sequence?}

The base action chunk could instead initialize rectified flow directly (\cref{tab:interface-diagnostics}(b)).
Removing the Mamba action encoder and flowing from the normalized base action chunk to the demonstration drops success to $29.00\%$, far below the unchanged base VLA ($72.56\%$) and \method ($74.44\%$).
The largest drops occur on Amount and Obstacle, where success falls to $5.67\%$ and $19.67\%$, respectively, and preserving sequential action structure while revising long or abruptly changing trajectories is critical.
Thus, the base action chunk is more effective as a persistent encoded condition than as generation's initial geometry alone.

% ===== END experiments/experiments_v7 =====
% ===== BEGIN conclusion/conclusion_v7 =====
\section{Conclusion}
\label{sec:conclusion}

We introduced \method, a base-VLA-guided world foundation model for moving-object manipulation.
\method regenerates action chunks from base action chunks, recent multi-view evidence, and robotic-arm states without accessing internal VLA representations or continuing to train the base VLA.
Across fine-tuned and coarse-tuned checkpoints of SmolVLA, X-VLA, $\pi_0$, and $\pi_{0.5}$, \method improves all eight model-level averages on \benchmark, with having performance gains of $1.88\%$--$45.31\%$.
The strongest improvements appear for base VLAs with weaker dynamic performance and for task families involving velocity variation and collision-induced path changes.
These results show that \method enhances the moving-object manipulation capability of fine-tuned and coarse-tuned base-VLA checkpoints.

% ===== END conclusion/conclusion_v7 =====

\begin{credits}
\subsubsection{\ackname}
This work was supported in part by Beijing Municipal Science \& Technology Commission No.~Z251100004625090 and by the National Science Foundation of China (NSFC) under Grant No.~62176134.

\subsubsection{\discintname}
The authors declare no competing interests.
\end{credits}

% ===== BEGIN embedded bibliography from build/main_v7_pricai.bbl =====

% ===== END embedded bibliography =====

\end{document}